\documentclass[conference]{IEEEtran}
\usepackage[utf8]{inputenc}

\usepackage{graphicx}
\usepackage{siunitx}
\usepackage{textgreek}
\usepackage{url,hyperref,lineno,microtype,subcaption}

\usepackage[acronym,shortcuts,nonumberlist,nopostdot]{glossaries}
\newacronym{hri}{HRI}{Human-Robot Interaction}
\newacronym{aa}{AA}{Adaptive Autonomy}
\newacronym{loa}{LoA}{Levels of Autonomy}
\newacronym{ros}{ROS}{Robot Operating System}
\newacronym{hai}{HAI}{Hospital-Acquired Infections}
\newacronym{uv}{UV}{Ultraviolet}
\newacronym{uvc}{UVC}{Ultraviolet-C}
\newacronym{fov}{FoV}{Field-of-View}
\newacronym{led}{LED}{Light-Emitting Diode}
\newacronym{vr}{VR}{Virtual Reality}
\newacronym{nat}{NAT}{Network Address Translation}
\newacronym{slam}{SLAM}{Simultaneous Localization and Mapping}
\newacronym{pir}{PIR}{Passive Infrared}
\newacronym{stun}{STUN}{Session Traversal Utilities for NAT}
\newacronym{turn}{TURN}{Traversal Using Relays around NAT}
\newacronym{trl}{TRL}{Technology Readiness Level}
\newacronym{webrtc}{WebRTC}{Web Real-Time Communication}

\begin{document}

\title{An Open-Source Modular Robotic System for Telepresence and Remote Disinfection}

\author{\IEEEauthorblockN{Andre Potenza}
\IEEEauthorblockA{\textit{Center for Applied Autonomous Sensor Systems (AASS)} \\
\textit{Örebro University}\\
Örebro, Sweden \\
andre.potenza@oru.se}
\and
\IEEEauthorblockN{Andrey Kiselev}
\IEEEauthorblockA{\textit{Center for Applied Autonomous Sensor Systems (AASS)} \\
\textit{Örebro University}\\
Örebro, Sweden \\
andrey.kiselev@oru.se}
\and
\IEEEauthorblockN{Alessandro Saffiotti}
\IEEEauthorblockA{\textit{Center for Applied Autonomous Sensor Systems (AASS)} \\
\textit{Örebro University}\\
Örebro, Sweden \\
asaffio@aass.oru.se}
\and
\IEEEauthorblockN{Amy Loutfi}
\IEEEauthorblockA{\textit{Center for Applied Autonomous Sensor Systems (AASS)} \\
\textit{Örebro University}\\
Örebro, Sweden \\
amy.loutfi@oru.se}
}

\maketitle

\begin{abstract}
In a pandemic contact between humans needs to be avoided wherever possible. Robots can take over an increasing number of tasks to protect people from being exposed to others. One such task is the disinfection of environments in which infection spread is particularly likely or bears increased risks. It has been shown that UVC light is effective in neutralizing a variety of pathogens, among others the virus causing COVID-19, SARS-CoV-2. Another function which can reduce the need for physical proximity between humans is interaction via telepresence, i.e., the remote embodiment of a person controlling the robot.

This work presents a modular mobile robot for telepresence and disinfection with UVC lamps. Both operation modes are supported by adaptable autonomy navigation features for facilitating efficient task execution. The platform’s primary contributions are its hardware and software design, which combine consumer-grade components and 3D-printed mountings with open-source software frameworks.

\end{abstract}

\section{Introduction}

The global COVID-19 pandemic has disrupted public life and entire industries. Exposing a severe lack of preparedness in the healthcare sector, among others, administrations scrambled to adjust to the new reality and prevent contagion in critical environments such as hospitals and retirement homes. While protective personal equipment was made available within weeks and cleaning protocols ramped up, more actions can be taken to avoid a spread of the virus and other pathogens. Mobile robots are now entering a stage in which they have the potential to make a difference in fighting the current and future pandemics~\cite{yang2020combating}.
Two types of measures can help protect both healthcare workers and patients, as well as older adults as the primary risk group. Telepresence (or telemedicine) can reduce the physical contact between healthcare workers and caregivers on the one side and patients on the other~\cite{yang2020keep}. Second, better hygiene in hospitals further reduces the risk of transmission, especially between patients.

Disinfection is an essential concern in hospital care. Even before the first appearance of COVID-19, recent years have seen a growing interest in alternative and supplementary methods for disinfecting surfaces in high-risk environments. Besides cost pressure, this development was precipitated by the increased occurrence of \ac{hai}, which are associated with a rise in morbidity and deaths across a large number of countries~\cite{barrasa2017impact}. \ac{hai}s are most frequently attributed to several types of bacteria, some of which over the years have developed resistances to a wide range of antibiotics~\cite{khan2015nosocomial}.
Standard cleaning procedures frequently fail to remove a substantial share of potentially harmful microbes on high-touch surfaces. As a result, \ac{hai}s have become a common cause of complications and deaths in patients that are immuno-compromised or recently underwent surgery. 

For decades, no-touch disinfection with \ac{uvc} light has been applied in water purification~\cite{song2016application}, as it decontaminates without the need for chemical disinfectants. As such, it represents a more environment-friendly and safer alternative. Only in recent years it has been identified as a viable option for decreasing the microbial burden on surfaces in hospitals, e.g., in patient and operating rooms~\cite{andersen2006comparison}. As an adjunct procedure to standard cleaning protocols using chemical cleaning agents it has been shown to further reduce the density of microbes on surfaces. Although the majority of \ac{hai}s are caused by bacteria, \ac{uvc} is similarly effective against viruses~\cite{sabino2020uv} and fungi. Furthermore, unlike many chemical sanitizers, it leaves no residues that might pose long-term health risks.

This work presents the design and implementation of an open-source modular disinfection and mobile robotic telepresence system for prototyping and research. The system provides functionality for environment mapping and semi-autonomous navigation. The two operation modes, \textit{telepresence} and \textit{\ac{uvc} disinfection}, rely on the same architecture and interface and only differ with regards to the hardware equipment on the robot (i.e., the lamps which take the place of the screen). 

While there are a variety of telepresence and a few disinfection robots available on the market, the main motivation behind the presented work is to offer an approach and methodology for building and deploying such robots on demand and within a narrow time frame. This implicates specific requirements, especially with respect to the hardware. The approach taken in the present work relies on off-the-shelf components, simple assembly, and affordable manufacturing methods. The software part of the project is designed for rapid deployment and engineered in a way that permits easy modification.

The remaining part of the paper is organized as follows. In Section \ref{sec:related_work} we discuss relevant work and developments related to both robotic telepresence and the use of \ac{uv} lamps for disinfection in hospitals, as well as the few efforts combining mobile robotics and \ac{uvc} disinfection. Section \ref{sec:implementation} provides a detailed description of hardware and software design of our platform. Subsection \ref{subsec:adaptivity} is concerned with the design and behavior of the robot's (semi-)autonomous functionality, followed by the conclusion in Section~\ref{sec:conclusion}.

\section{Related Work}
\label{sec:related_work}

\subsection{Robotic Telepresence}
In a minimal configuration, a telepresence robot possesses a mobile base, which can be controlled remotely to navigate through its environment, and a device for telecommunication, e.g., a tablet, smartphone or standard computer, along with a wireless internet connection. In order to facilitate 'face-to-face' interaction, the screen and camera are typically mounted in an elevated position on top of the base~\cite{Kristoffersson2013}. 
In a telepresence interaction the person controlling the robot is often referred to as the \textit{remote user} or \textit{operator}, whereas the \textit{local user(s)} are located in the same physical environment as the robot. Unlike standard communication devices, robotic telepresence affords the user a certain degree of freedom to explore and interact with the local environment independently. 

In addition, with the introduction of advanced features such as adaptable autonomy, users can be supported in controlling the robot efficiently while focusing on relevant tasks~\cite{potenza2017towards}.
Common application domains for telepresence robots include elderly care~\cite{orlandini2016excite, Beer2011}, healthcare (as a form of telemedicine)~\cite{garingo2016tele}, as well as telecommuting in office or industrial environments~\cite{Tsui2011a}.

The range of action of current telepresence platforms is still fairly limited. Besides the common limitations related to wheel-based locomotion and a lack of (effective) manipulators, their performance often suffers from a reduced sense of remote presence and awareness within the robot's environment. Important contributing factors to this impairment are the often insufficient sensor resolution and \ac{fov} of common webcams~\cite{johnson2015can, Kiselev2014}. One of the goals in designing a telepresence robot is to enhance the user experience and control performance by providing visual and auditory information that matches as closely as possible people's natural perception of the world, either by using a camera with a wide angle or mounting it on an articulated element, so as to simulate head movements~\cite{karimi2018mavi}.

Various platforms with different sets of features are available as consumer products on the market. In recent years it has become increasingly common to provide assisted driving and other convenient capabilities beyond basic teleoperation and telecommunication (e.g., Double 3\footnote{https://www.doublerobotics.com} and Ava\footnote{www.avarobotics.com}). However, such commercial products are generally not designed to be customizable, as they do not provide access to components and utilize proprietary software.

\subsection{Disinfection with UVC Light}
With a wavelength spectrum ranging from approximately 100 to \SI{280}{\nano\meter}, \ac{uvc} light from the Sun is absorbed entirely by the ozone layer. As a consequence, the only sources on the Earth's surface are artificial. \ac{uvc} light causes inactivation of microorganisms by damaging the RNA and DNA, thereby inhibiting their reproduction~\cite{cutler2011ultraviolet}. Due to its strong germicidal properties it is well suited for disinfection in a wide range of scenarios.

The time required for disinfecting a surface depends on several factors. The maximal dose of light absorbed by a surface is a function of the distance from the \ac{uvc} source and the output of the source, and is cumulative over time. The dose is measured in \SI{}{\joule/m^2} (in practice often in \SI{}{\micro\watt\second\per\centi\meter^2})~\cite{andersen2006comparison}. The surface material and presence of organic matter can play a role in the effectiveness of the procedure. The required dose for inactivation further depends on the initially present microbial load and can vary between types of microorganisms.
In clinical tests, the effectiveness of a disinfection process is commonly expressed in log reductions of a known initial quantity of pathogens on a test surface as a result of exposure from a reference distance and over a fixed amount of time~\cite{yang2019effectiveness}.

Today, the most widely used devices are fluorescent lamps based on low-pressure mercury vapor and emit \ac{uvc} light at a peak wavelength of \SI{253.7}{\nano\meter}. Aside from the ultraviolet radiation, they give off visible light of a blueish-white hue. Other types include pulsed xenon lamps~\cite{simmons2020deactivation, nerandzic2015evaluation} and in recent years \ac{uv} \ac{led} lamps have been shown to produce similar results in inactivating bacteria and viruses~\cite{beck2017evaluating, nyangaresi2019comparison}. \ac{led} lamps are especially promising for deployment on robots, as they are significantly more energy-efficient than fluorescent lamps and easier to install. However, at the time of writing they are not yet available in adequate and certified devices.

Early studies concerned with evaluating the efficacy of \ac{uvc} light for disinfection mainly sought to compare it with conventional methods, i.e., chemical disinfection~\cite{havill2012comparison}. The lamps used in these studies were deployed manually and moved between cycles. In some more recent efforts targeting practical use, arrays of lamps were mounted on a passive wheeled platform to be moved manually from one position to another~\cite{yang2019effectiveness}.
By placing several petri dishes with test cultures in different locations in a hospital room it is possible to obtain evidence of efficacy under realistic conditions.

Despite the documented conducive effect of disinfection with \ac{uvc} light, it has thus far only found sparse adoption in the healthcare sector. This is attributable to a variety of factors, not least a still unfavorable cost-benefit equation. Moreover, deployment often remains impractical and associated with added labor rather than an easing of the burden on healthcare and custodial staff.

One drawback of \ac{uvc} light is that it is harmful to the skin and eyes of higher organisms, including humans, which is why exposure must be strictly avoided. Therefore, robots lend themselves as an obvious alternative for operating \ac{uvc} emitting devices and reducing human involvement to the extent possible. Nevertheless, staff operating the lamps need to be trained in handling them safely to avoid inadvertent exposure. 

\subsubsection{Practical Limitations}
In practice, the efficacy of \ac{uvc} light as a disinfectant is limited by natural constraints, such as the need for direct surface exposure. That is, any occluded surfaces receive a significantly smaller dose, which is too low for complete disinfection. Proximity of the light source to the target surface is crucial, as intensity decreases in proportion to the inverse square of its distance from that surface. In addition, since \ac{uvc} is absorbed by most materials, its effect is diminished in the presence of organic residues such as blood or urine~\cite{andersen2006comparison}.

Although extensible robotic arms could potentially reach surfaces that are otherwise shadowed and UV-reflective wall coating has been shown to reduce the duration of disinfection cycles~\cite{rutala2013rapid}, neither method can guarantee that all sites receive a sufficient dose. As a consequence, \ac{uvc} disinfection is inadequate as a standalone cleaning procedure. Rather, it can complement environmental and spot cleaning as an additional layer towards further decreasing the risk of nosocomial infections.

\section{System Architecture and Implementation}
\label{sec:implementation}
A major impediment in the early stages of the pandemic were intellectual property regulations preventing rapid large-scale production of critical equipment. Examples of such products range from simple adapters to complete ventilator designs. Vital products had to be redesigned and released under various open-source licences to allow manufacturing on demand. 
Our main purpose in developing the presented system was to make it available as an open-source project that can be implemented by anyone. Therefore, the focus was placed on simplicity, ease of manufacturing, as well as software and hardware licensing. 

Another central factor shaping the overall architectural design of the robotic system is its adaptability to a wide variety of user requirements. Being primarily intended as a research platform for experimentation in robotic telepresence, modularity and adaptability are vital both with respect to hardware and software. At the same time, the development was also motivated by the intention to make the system available to a wider community, as a platform for implementing robotic solutions for telepresence and remote disinfection. 

\begin{figure}[t]
    \begin{center}
        \includegraphics[width=\linewidth]{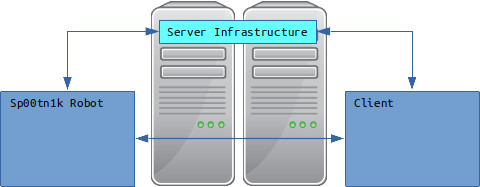}
    \end{center}
    \caption{High-level view of the system architecture in telepresence mode.}
    \label{fig:sp00tn1k-architecture-hl}
\end{figure}

An overview of the system's three major components is shown in Fig.~\ref{fig:sp00tn1k-architecture-hl}. They include the robotic platform (Sp00tn1k Robot), its control interface (Client), and the supporting server for signaling and \ac{nat} traversal (Server Infrastructure) \footnote{The source code of all components and blueprints are available at https://bitbucket.org/sp00tn1k}. The latter two parts are not required if the final system does not foresee telepresence or remote control features.

\subsection{Robot Hardware Design}
\label{subsec:hardware}
The hardware design is based on an open-frame approach which leaves structural components of the robot exposed. It uses slotted aluminum extrusions to simplify assembly and allow flexibility in mounting components. For the most part, the hardware design makes use of standard consumer-grade off-the-shelf components to allow rapid customization, assembly, and deployment in different geographical locations. 

\begin{figure*}[t]
    \centering
    \includegraphics[width=0.9\textwidth]{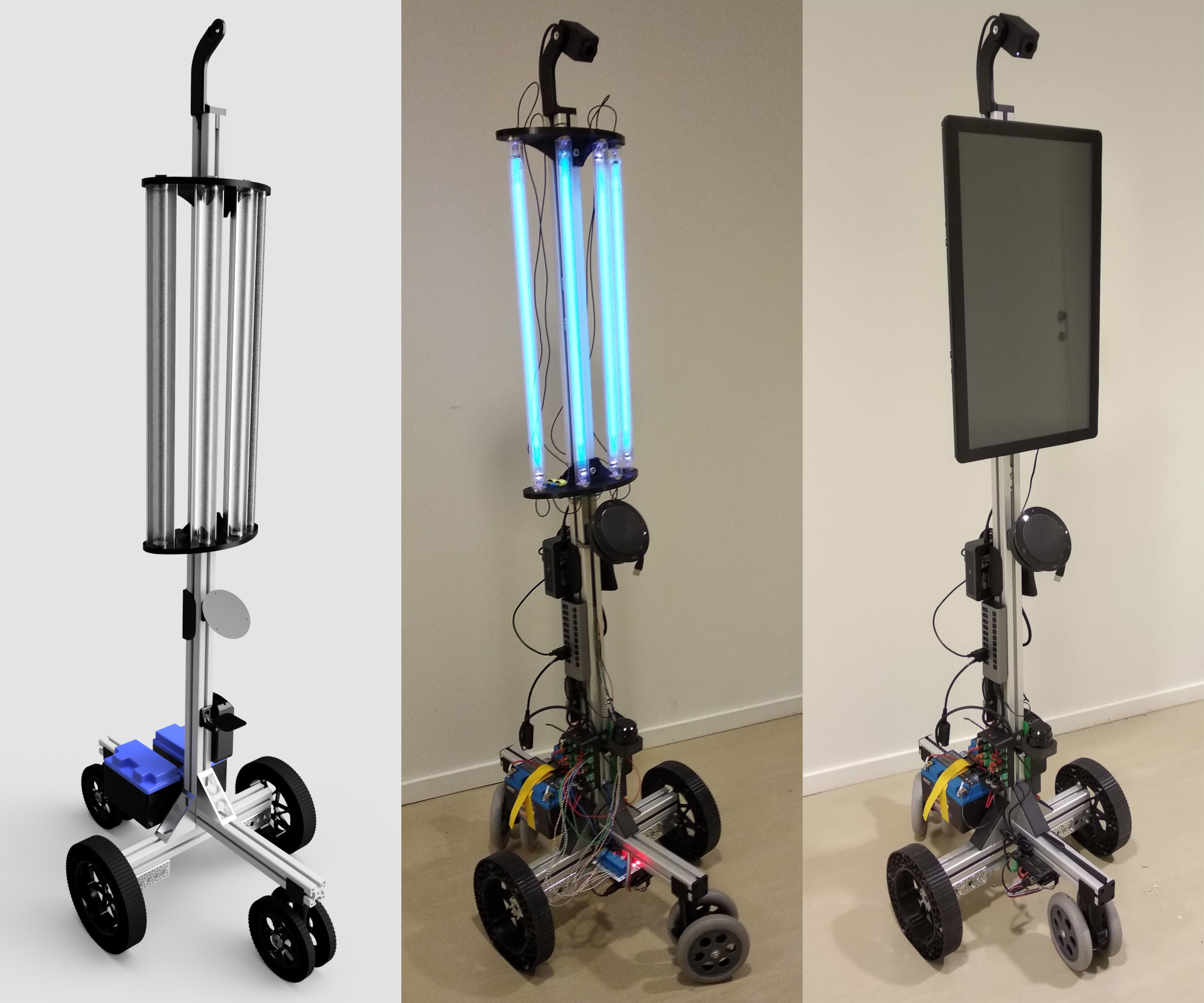}
    \caption{Sp00tn1k robot. Left: CAD assembly view with mounted UVC lamps. Center: Assembled robot. Right: Robot with display instead of lamps.}
    \label{fig:sp00tn1k-assembly}
\end{figure*}

As is common for telepresence robots, the platform consists of a wheel base and a screen mounted on a pole on top of the base to provide the various functionalities. It is further equipped with a 2D LIDAR for \ac{slam} and navigation, as well as a wide angle camera for operation with an extended \ac{fov}. The wheels and mounts for sensors and electronics as well as casings are printed from PLA material. 
The robot was designed as a hybrid to be used for telepresence and \ac{uvc} disinfection, with the screen and \ac{uv} lamps being swapped out between the two modes of operation (as seen in Fig.~\ref{fig:sp00tn1k-assembly}). 

Below, we describe the various components and infrastructure in more detail.

\begin{figure*}[ht]
    \centering
        \includegraphics[width=0.9\linewidth]{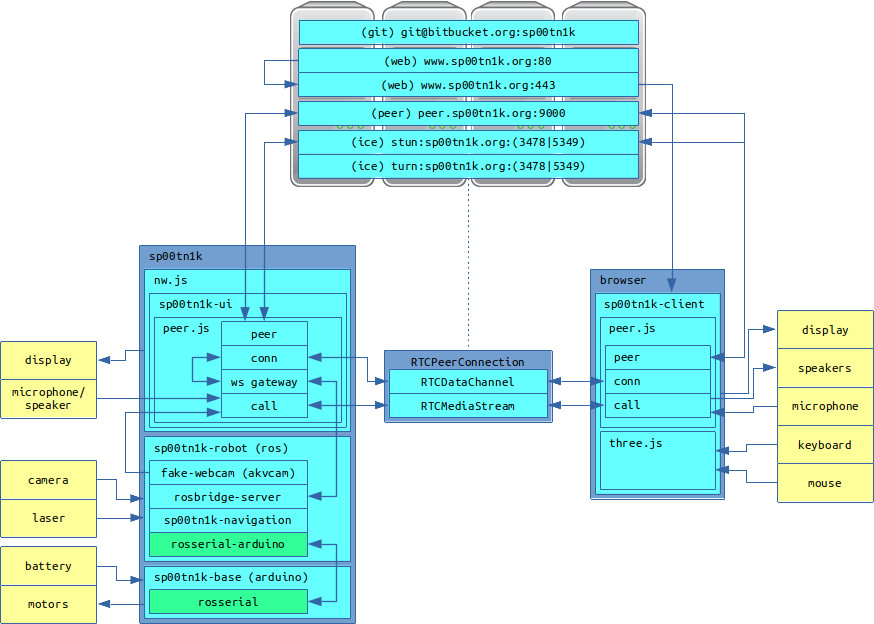}
    \caption{Overall system architecture in telepresence mode}
    \label{fig:sp00tn1k-architecture}
\end{figure*}

\begin{figure*}[t]
    \centering
        \includegraphics[width=0.9\textwidth]{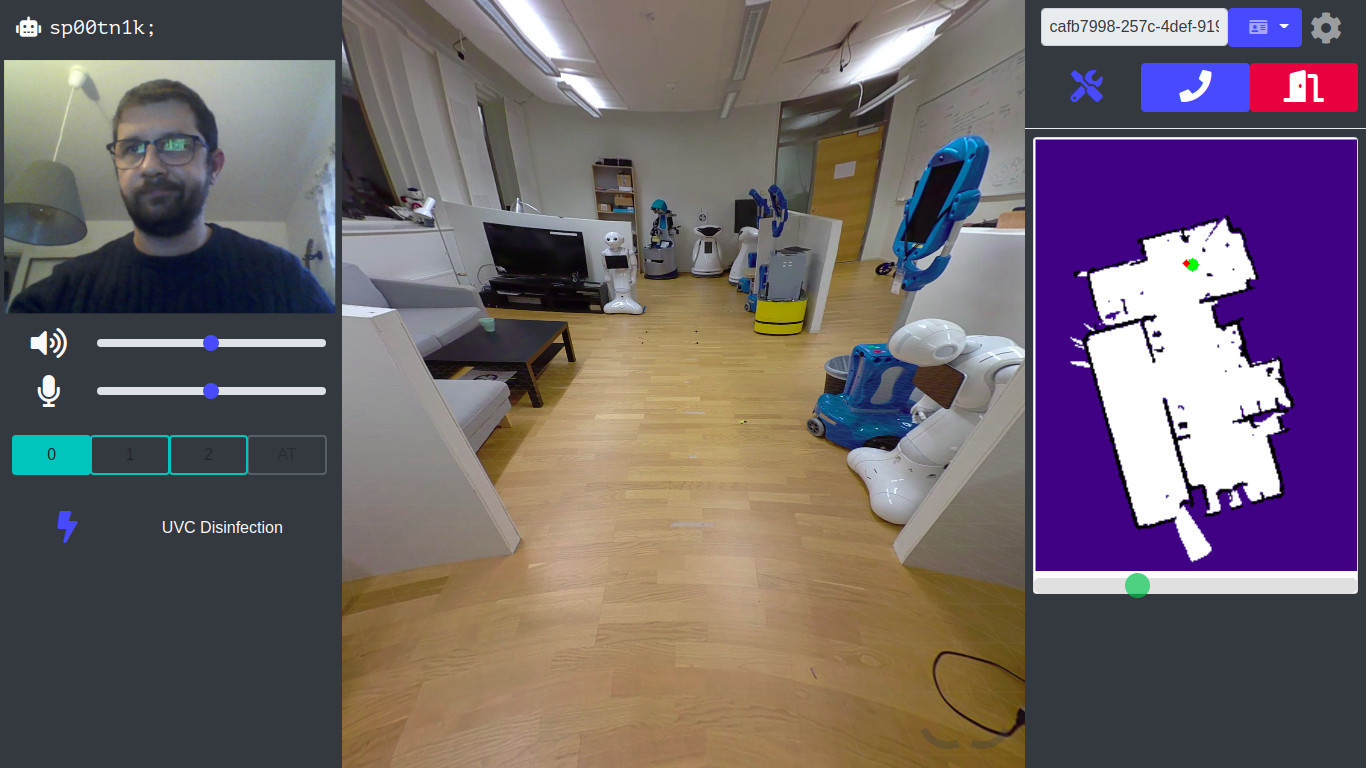}
    \caption{The client interface. Left: The remote user's own video stream, autonomy level slider and disinfection lamp control. Center: The robot's video stream. Right: A map of the environment.}
    \label{fig:sp00tn1k-interface}
\end{figure*}

\subsubsection{Chassis and Wheels}
The frame consists of T-Slot 40 aluminum extrusions which render the hardware platform highly modular and reconfigurable. The robot's kinematics consist in a symmetrical differential drive, which allows rotating in place. The two motor-driven wheels on the sides are supported by a caster wheel in the front and rear each, resulting in an overall footprint of 49x62cm (WxL). 
The vertical pole is attached centrally on top of the base. This simple design provides high stability, capable of supporting relatively large payloads such as a 22" touchscreen display at a high mounting point. Additionally, universal mounting adapters can be attached and rearranged freely on the T-Slots. Electronic components are mounted onto the frame with cabling hidden inside the extrusion where possible (see Fig.~\ref{fig:sp00tn1k-assembly}). 
While this design yields a high degree of flexibility, it also exposes electrical connections, which can be undesirable in a number of cases. With the camera sitting on top of the pole, the robot measures \SI{1.65}{m} in height. It weighs \SI{17.5}{kg} in the telepresence configuration and \SI{13.6}{kg} with the disinfection lamps mounted. 

The base does not have any built-in suspension. Instead, the flexible wheels enable the robot to traverse small bumps, such as carpets and thresholds. This design is prone to higher instabilities in comparison with traditional suspension with dampers and stabilizer bars, but usually performs better than inflated wheels. The wheels are 3D-printed in two parts, using PLA material for the rims and TPU 95A for the tires.
The robot is powered by two parallel 12V lithium-iron-phosphate (LiFEPo4) batteries with a capacity of \SI{60}{Wh} each. 
Under normal load the average runtime is roughly 3 hours.
The motors are Actobotics \SI{12}{V} 313RPM DC planetary gear motors with a 27:1 gear ratio. 
The low-level controller runs on an Arduino Mega 2560 board and uses a Cytron 10A 7-30V dual channel motor driver. 

\subsubsection{Components and Sensors}
The robot’s computer is an Intel® NUC 10 (NUC10i7FNK) with 32GB RAM and a solid state drive. The screen for the telepresence mode is a 22" iiyama ProLite TF2215MC-B2 which is rotated in portrait mode. In addition, a Jabra SPEAK 410 speakerphone provides high quality audio input and output, while a Huddly IQ camera delivers the video stream from the local environment.
A 360$^{\circ}$ 2D laser range finder (RPLIDAR S1) detects obstacles on a plane around the robot. This information is used for several processes, including mapping, localization, autonomous navigation and assisted teleoperation.

\subsubsection{Disinfection Lamps}
In the current setup the platform is equipped with 4 fluorescent \ac{uvc} lamps (Philips T8, \SI{16.7}{W}) with a length of \SI{590}{mm}. The lamps are attached on the pole at a height of approximately \SI{1.2}{m} and arranged in a semicircle in the front. The \ac{uvc} output per lamp is equal to \SI{4.5}{W}, which corresponds to a maximum intensity of \SI{35.8}{\micro\watt\per\centi\meter^2} 
at 1m distance. 
For efficient use in real-world scenarios a higher wattage as well as possibly additional lamps are highly suggested.
The lamps are controlled through the client and set to be deactivated when the connection with the client is lost.

\subsubsection{Modularity}
Since the screen and lamps occupy the same space on the robot, they need to be replaced when switching between the operation modes.
The components listed above are an example configuration. In most cases there are less expensive equivalents that serve the purpose sufficiently well. Individual components can be omitted altogether, depending on the application.

\subsection{Robot Software Design}
\label{subsec:software}
The two major components of the software infrastructure are \ac{ros} and \ac{webrtc}. Being fully open-source, they offer the necessary interfaces both on a process level on the robot and for web-based communication with and through the client. 
\ac{webrtc} was selected as a means of communication between the robot and its control interface. The main motivation for using \ac{webrtc} is its wide support by most modern web browsers, as well as its ability to fully handle communication between peers via media and data channels. It further supports open-source codecs for media streams, such as VP8 and OPUS, and implements security on the communication channel level. As a result, the system can be controlled using any device capable of running modern browsers with \ac{webrtc} support.

Developers are only required to implement signalling between peers and handle media and data streams on the robot and control interface. On sp00tn1k the PeerJS library~\cite{peerjs} is responsible for these tasks. 

\subsubsection{ROS Infrastructure}
The platform itself runs on the open-source \ac{ros} framework (noetic distribution on Ubuntu 20.04), making use of readily available and established packages such as the \ac{ros} navigation stack, \textit{rosserial} and \textit{cartographer} for \ac{slam}. In addition, several packages were developed to provide the back end functionality of the client UI, as well as the assisted teleoperation mode. The \ac{ros} package \textit{rosbridge server} allows transmission of \ac{ros} messages to and from the client through \ac{webrtc}'s data channel. 

\subsubsection{Client and Robot UI}
The client interface can be controlled via mouse and touch input (see Fig.~\ref{fig:sp00tn1k-interface}). Located in the center of the screen is the video stream from the robot's camera showing the local environment. Two sidebars to the left and right host the client's own video stream, a map of the environment (if available) and further inputs for the robot's controls, e.g., for activating and deactivating the \ac{uvc} lamps. The video stream is embedded in a 3D-rendered environment. This allows displaying other relevant information, as well as switching to a \ac{vr} mode.
New features can be introduced via a plugin system without interfering with the rest of the infrastructure.

The interface on the robot's side is minimal, comprising primarily the incoming video stream from the client's side. Like the client, it can be extended to host additional information and functionality, such as a local control mode.

\subsubsection{Communication}
The robot connects to a peering server on startup and waits for incoming connections. The client software is loaded from the web server, and also connects to a peering server when loaded. Subsequently, the client can attempt to connect and log on to the robot with a known ID. The \ac{stun} and \ac{turn} servers are required to establish a direct connection between the robot and a client.

Two channels mediate communication between the client and the robot. The RTCMediaStream transmits full-duplex audio and video, and the RTCDataChannel is responsible for control commands and telemetry data. Both are contained in an RTCPeerConnection object (see Fig.~\ref{fig:sp00tn1k-architecture}). Because the robot uses \ac{ros} as its control framework, the client software is also "\ac{ros}-aware", i.e., it communicates with the robot through \ac{ros} messages. 

On the robot side, the audio stream is routed directly to its audio devices. The incoming video stream is displayed in the UI and the outgoing stream is supplied by a virtual loopback webcam device. This way the camera image can be preprocessed before being streamed to the client.

Control messages are transparently passed by the robot's communication part (\textit{ws gateway} on the diagram) to \ac{ros} via the local WebSocket connector. Navigation messages arriving from the client at the \textit{rosbridge} server are passed on to the driver controller by means of \textit{rosserial} communication.

\subsection{Adaptable Autonomy}
\label{subsec:adaptivity}
In order to facilitate multi-purpose deployment and satisfy different user requirements and preferences, the platform can navigate at various degrees of autonomy. By clicking on a position on the map, be it static (prerecorded) or dynamic (\ac{slam}), the robot leverages the \ac{ros} navigation stack to plot and follow a route to the appropriate position in the environment. The robot is teleoperated manually by clicking on the central video image. A disk-shaped element projected on the floor indicates the approximate location to where the robot will move. Teleoperation can be used either without any support or with assistance. This entails collision avoidance by decelerating in close proximity to nearby objects. Moreover, in a second assistance mode, the robot steers away from obstacles to either side. The interface further affords seamless switching between the different navigation modes, thus allowing users to quickly adapt to new situations.

Two of the navigation modes (assisted teleoperation and autonomous) were evaluated in a previous user study~\cite{olatunji2020levels}. Following the study results, the client's responsiveness and feedback were augmented to offer enhanced transparency when operating in the assisted or autonomous mode. A central aspect in the design of autonomous functions is to ensure that users maintain a high degree of situation awareness while not incurring excess mental workload~\cite{beer2014toward}. To prevent confusion and frustration, a GUI element located in the sidebar of the interface displays which mode is active at any moment (see Fig.~\ref{fig:sp00tn1k-interface}).

\section{Conclusion}
\label{sec:conclusion}
The platform described in this paper is a research-oriented work in progress and as such not ready to be deployed for disinfection. However, we hope that it can serve as a basis for the development of fully operational devices.

It is believed that pandemics like the current one are going to become more frequent and we understand now that we need to be better prepared for the next one. It is likely that both mobile robotics and disinfection with \ac{uvc} light will play a vital role in combating future outbreaks and preventing widespread infection with contagious diseases in public spaces.

Even though the utility of most available mobile robots is still fairly limited, telepresence robots may, in the not too distant future, become true embodiments of people in the world. 

\subsection{Limitations and Future Work}
In its present form the robot cannot mechanically manipulate its surroundings. As a consequence, autonomous disinfection relies on appropriate infrastructure or provisions in the environment where it is deployed. Doors and stairs represent barriers that need to be accounted for. 
As previously noted, the \ac{uvc} output of the current configuration is relatively low. A higher wattage increases the dose delivered to a nearby surface in a given time and thus reduces the time the robot needs to stop at each visited location accordingly. In turn, a higher battery capacity can compensate for the increased energy intake to yield a reasonable runtime between charge cycles.

An integrated task planner will allow users to plan and schedule disinfection in a larger area (e.g., an entire hospital wing) from within the client. At a subsequent stage the robot might be taught to select appropriate poses for disinfecting a provided selection of prioritized surfaces and sites based on respective distances and angles. This way the need for human intervention could be further reduced.

The assisted teleoperation currently relies on rule-based methods to avoid or soften collisions with the environment. In a next step machine learning may further enhance the capacity to provide assistance by leveraging feedback from the user and learning their preferences.

With the robot conceivably being deployed in a variety of different environments and scenarios, local users should be enabled to control it as well as the remote user. For this purpose, different interfaces may qualify, such as game controllers, the touch screen or even speech and gestures. Yet another useful feature both for remote and local control is autonomous people following~\cite{olatunji2020user}.

A key aspect for the safe use and acceptance of telepresence robots is privacy. As usage scenarios vary with respect to the scenario and overall purpose, so do usage and access rights – both of remote and local users. In a public environment such as a hospital, there may be telepresence platforms for anyone to log in to. In private homes, however, it would be up to the local user to accept a requested login, analogous to any conventional telecommunication device. Yet other arrangements are conceivable if a caregiver is to check in regularly on a patient or relative. Accordingly, any commercial platform should be able to handle these different scenarios.

\section*{Funding}
This work has received funding from the European Union’s Horizon 2020 research and innovation program under the Marie Skłodowska-Curie grant agreement No 721619 for the SOCRATES project and the testbed for Autonomous Intelligent Machines for Enterprise and Exploration (AI.MEE) at Örebro University.

\section*{Acknowledgments}
The authors express gratitude to Samer Dia for his work on the design and manufacturing of the flexible wheels.

\IEEEtriggeratref{21}
\bibliographystyle{ieeetr} 
\bibliography{library}

\end{document}